\documentclass{article}

\usepackage{microtype}
\usepackage{graphicx}
\usepackage{subfigure}
\usepackage{booktabs} 

\usepackage{xspace}
\makeatletter
\DeclareRobustCommand\onedot{\futurelet\@let@token\@onedot}
\def\@onedot{\ifx\@let@token.\else.\null\fi\xspace}
\def\eg{\emph{e.g}\onedot} 
\def\ie{\emph{i.e}\onedot} 
 
\def\etc{\emph{etc}\onedot}

\makeatother

\usepackage{color}

\usepackage{amssymb}
\usepackage{amsmath,amsfonts,bm}

\usepackage{verbatim}
\usepackage{wrapfig}
\usepackage{graphicx}

\usepackage{hyperref}

\usepackage[accepted]{icml2021}

\icmltitlerunning{L2E: Learning to Exploit Your Opponent}

\begin{document}

\twocolumn[
\icmltitle{L2E: Learning to Exploit Your Opponent}

\icmlsetsymbol{equal}{*}

\begin{icmlauthorlist}
\icmlauthor{Zhe Wu}{equal,to1,to2}
\icmlauthor{Kai Li}{equal,to1,to2}
\icmlauthor{Enmin Zhao}{to1,to2}
\icmlauthor{Hang Xu}{to1,to2}
\icmlauthor{Meng Zhang}{to1,to2}
\icmlauthor{Haobo Fu}{to3}
\icmlauthor{Bo An}{to4}
\icmlauthor{Junliang Xing}{to1,to2}
\end{icmlauthorlist}

\icmlaffiliation{to1}{Center for Research on Intelligent System and Engineering, Institute of Automation, Chinese Academy of Sciences, China}
\icmlaffiliation{to2}{School of Artificial Intelligence, University of Chinese Academy of Sciences, China}
\icmlaffiliation{to3}{Tencent AI Lab, China}
\icmlaffiliation{to4}{School of Computer Science and Engineering, Nanyang Technological University, Singapore}

\icmlcorrespondingauthor{Kai Li}{kai.li@ia.ac.cn}
\icmlcorrespondingauthor{Junliang Xing}{jlxing@nlpr.ia.ac.cn}

\icmlkeywords{Reinforcement Learning, Opponent Modeling}

\vskip 0.3in
]

\printAffiliationsAndNotice{\icmlEqualContribution}

\begin{abstract}
Opponent modeling is essential to exploit sub-optimal opponents in strategic interactions.
Most previous works focus on building \emph{explicit} models to directly predict the opponents' styles or strategies, which require a large amount of data to train the model and lack adaptability to unknown opponents.
In this work, we propose a novel Learning to Exploit (L2E) framework for \emph{implicit} opponent modeling. L2E acquires the ability to exploit opponents by a few interactions with different opponents during training, thus can adapt to new opponents with unknown styles during testing quickly.
We propose a novel opponent strategy generation algorithm that produces effective opponents for training automatically.
We evaluate L2E on two poker games and one grid soccer game, which are the commonly used benchmarks for opponent modeling.
Comprehensive experimental results indicate that L2E quickly adapts to diverse styles of unknown opponents.
\end{abstract}

\section{Introduction}
\label{intro}
One core research topic in modern artificial intelligence is creating agents that can interact effectively with their opponents in different scenarios.
To achieve this goal, the agents should have the ability to reason about their opponents' behaviors, goals, and beliefs.
Opponent modeling, which constructs the opponents' models, has been extensively studied in past decades~\cite{albrecht2018autonomous}.
In general, an opponent model is a function that takes some interaction history as input and
predicts some property of interest of the opponent.
Specifically, the interaction history may contain the past actions that the opponent took in various situations, and the properties of interest could be the actions that the opponent may take in the future, the style of the opponent (\eg, ``defensive'', ``aggressive''), or its current goals.
The resulting opponent model can inform the agent's decision-making by incorporating the model's predictions in its planning procedure to optimize its interactions with the opponent.
Opponent modeling has already been used in many practical applications, such as dialogue systems~\cite{grosz1986attention}, intelligent tutor systems~\cite{mccalla2000active}, and security systems~\cite{jarvis2005identifying}.

Many opponent modeling algorithms vary greatly in their underlying assumptions and methodology.
For example, policy reconstruction based methods~\cite{powers2005learning,banerjee2007reaching} explicitly fit an opponent model to reflect the opponent's observed behaviors.
Type reasoning based methods~\cite{dekel2004learning,nachbar2005beliefs} reuse pre-learned models of several known opponents by finding the one that most resembles the current opponent's behavior.
Classification based methods~\cite{huynh2006integrated,sukthankar2007policy} build models that predict the opponent's play style, and employ the counter-strategy, which is effective against that particular style.
Some recent works combine opponent modeling with deep learning methods or reinforcement learning methods and propose many related algorithms~\cite{he2016opponent,foerster2017learning,wen2019probabilistic}.
Although these algorithms have achieved some success, they also have two obvious disadvantages:
1) constructing accurate opponent models requires a lot of data, which is problematic since the agent may not have the time or opportunity to collect enough data about its opponent in most applications;
 and 2) most of these algorithms perform well only when the opponents during testing are similar to the ones used for training, thus it is difficult for them to adapt to opponents with new styles quickly.

\begin{figure}[t]
	\centering
	\includegraphics[width=1\linewidth]{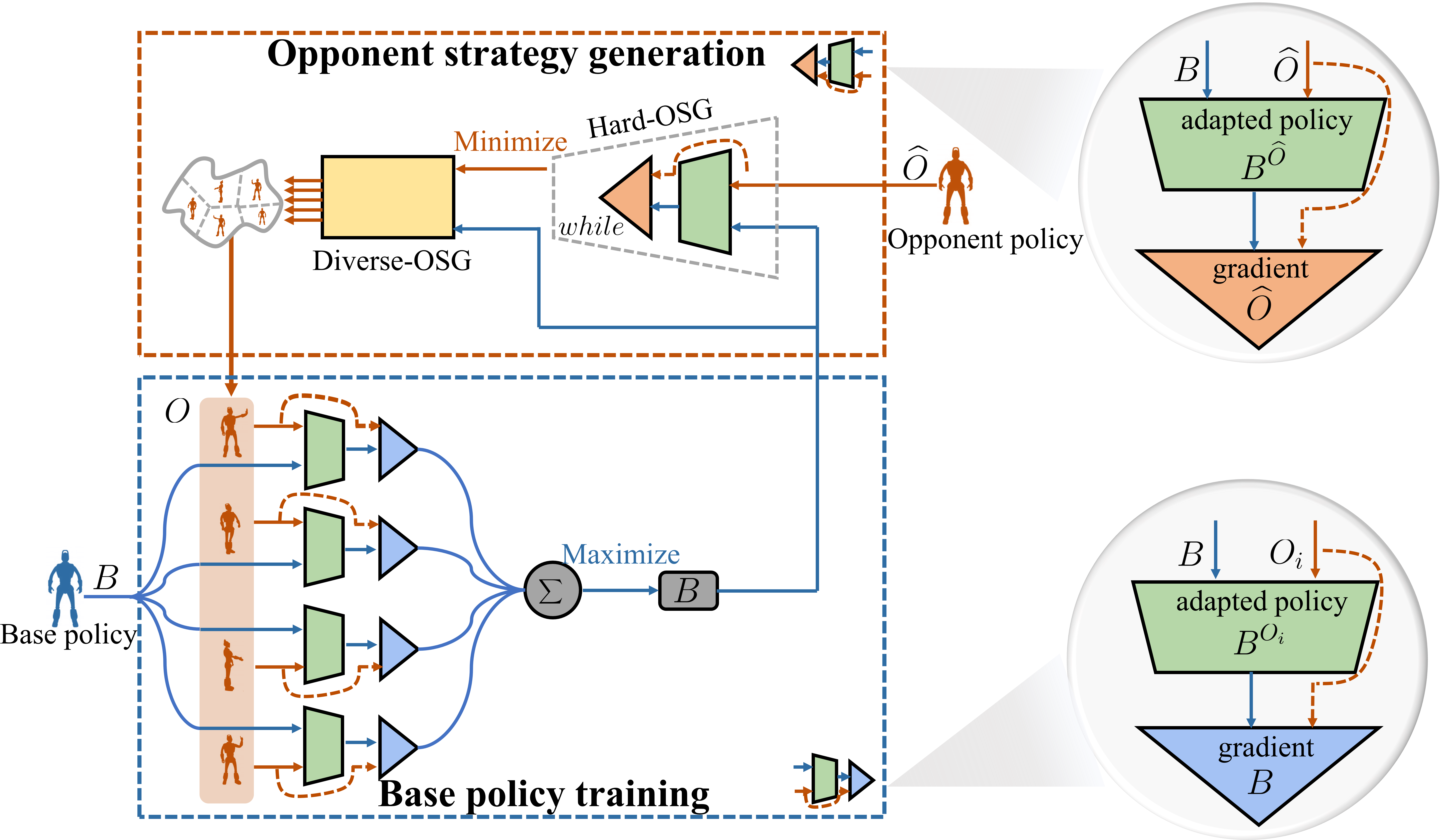}
	\caption{The overview of our proposed L2E framework.
		The base policy training part maximizes the base policy's adaptability by continually interacting with opponents of different strengths and styles (Section~\ref{Base Policy Training}). The opponent strategy generation part first generates hard-to-exploit opponents for the current base policy (Hard-OSG, see Section~\ref{Hard-to-Exploit Opponents Generation}), then generates diverse opponent policies to improve the generalization ability of the base policy (Diverse-OSG, see Section~\ref{Diverse Opponents Generation}). The resulting base policy can fast adapt to completely new opponents with a few interactions.}
	\label{overview}
	\vspace{-5mm}
\end{figure}

To overcome these shortcomings, we propose a novel Learning to Exploit (L2E) framework in this work for \emph{implicit} opponent modeling, which has two desirable advantages.
First, L2E does not build an explicit model for the opponent, so it does not require a large amount of interactive data and simultaneously eliminates the modeling errors.
Second, L2E can quickly adapt to new opponents with unknown styles, with only a few interactions with them.
The key idea underlying L2E is to train a \emph{base policy} against various opponents' styles by using only a few interactions between them during training, such that it acquires the ability to exploit different opponents quickly.
After training, the base policy can quickly adapt to new opponents using only a few interactions during testing.
In effect, our L2E framework optimizes for a base policy that is easy and fast to adapt.
It can be seen as a particular case of learning to learn, \ie, meta-learning~\cite{finn2017model}.
The meta-learning algorithm, such as MAML~\cite{finn2017model}, is initially designed for single-agent environments.
It requires the manual design of training tasks, and the final performance largely depends on the user-specified training task distribution.
The L2E framework is designed explicitly for multi-agent competitive environments, which automatically generates effective training tasks (opponents).
Some recent works have also initially used meta-learning for opponent modeling.
Unlike these works, which either use meta-learning to predict the opponent's behaviors~\cite{rabinowitz2018machine} or handle the non-stationary problem in multi-agent reinforcement learning~\cite{al2017continuous}, we focus on improving the agent's ability to adapt to unknown opponents quickly.

In our L2E framework, the base policy is explicitly trained such that a few interactions with a new opponent will produce an \emph{opponent-specific} policy to exploit this opponent effectively, \ie, the base policy has strong adaptability that is broadly adaptive to many opponents.
Specifically, if a deep neural network models the base policy, then the opponent-specific policy can be obtained by fine-tuning the parameters of the base policy's network using the new interactive data with the opponent.
A critical step in L2E is how to generate effective opponents to train the base policy.
The ideal training opponents should satisfy the following two desiderata.
1) The opponents need to be challenging enough (\ie, hard to exploit). By learning to exploit these challenging opponents, the base policy eliminates the weakness in its adaptability and learns a more robust strategy.
2) The opponents need to have enough diversity. The more diverse the opponents during training, the stronger the base policy's generalization ability is, and the more adaptable the base policy to the new opponents.

To this end, we propose a novel opponent strategy generation (OSG) algorithm, which can produce challenging and diverse opponents automatically.
We use the idea of adversarial training to generate challenging opponents.
Some previous works have also been proposed to obtain more robust policies through adversarial training and improve generalization~\cite{pinto2017robust,pattanaik2017robust}.
From the perspective of the base policy, giving an opponent, the base policy first adjusts itself to obtain an adapted policy, the base policy is then optimized to \emph{maximize} the rewards that the adapted policy gets when facing the opponent.
The challenging opponents are then adversarially generated by \emph{minimizing} the base policy's adaptability by automatically generating difficult to exploit opponents.
These hard-to-exploit opponents are trained such that even if the base policy adapts to them, the adapted base policy cannot take advantage of them.
Besides, our OSG algorithm can further produce diverse training opponents with a novel diversity-regularized policy optimization procedure.
In specific, we use the Maximum Mean Discrepancy (MMD) metric~\cite{gretton2007kernel} to evaluate the differences between policies. The MMD metric is then incorporated as a regularization term into the policy optimization process to obtain a diverse set of opponent policies.
By training with these challenging and diverse training opponents, the robustness and generalization ability of our L2E framework are significantly improved.
To summarize, this work's main contributions are listed below in three-fold:
\begin{itemize}
	\item We propose a novel learning to exploit (L2E) framework to exploit sub-optimal opponents without building explicit models for it. L2E can quickly adapt to unknown opponents using only a few interactions.
	\item We propose to use an adversarial training procedure to generate challenging opponents automatically. These hard to exploit opponents help L2E eliminate the weakness in its adaptability effectively.
	\item We further propose a diversity-regularized policy optimization procedure to generate diverse opponents automatically. L2E's generalization ability is improved significantly by training with these diverse opponents.
\end{itemize}
We conduct extensive experiments to evaluate the L2E framework in three different environments. Comprehensive experimental results demonstrate that the base policy trained with L2E quickly exploits a wide range of opponents compared to other algorithms.

\section{Related Work}
\subsection{Opponent Modeling}
\label{opp_modeling}
Opponent modeling is a long-standing research topic in artificial intelligence, and some of the earliest works go back to the early days of game theory research~\cite{brown1951iterative}.
The main goal of opponent modeling is to interact more effectively with other agents by building models to reason about their intentions, predicting their next moves or other properties~\cite{albrecht2018autonomous}.
The commonly used opponent modeling methods can be roughly divided into policy reconstruction, classification, type-based reasoning, and recursive reasoning.
Policy reconstruction methods~\cite{mealing2015opponent} reconstruct the opponents' decision-making process by building models that make explicit predictions about their actions.
Classification methods~\cite{weber2009data,synnaeve2011bayesian} produce models that assign class labels (\eg, ``aggressive'' or ``defensive'') to the opponent and employ a precomputed strategy that is effective against that particular class of opponent.
Type-based reasoning methods~\cite{he2016opponent,albrecht2017reasoning} assume that the opponent has one of several known types and update the belief using the new observations obtained during the real-time interactions.
Recursive reasoning methods~\cite{muise2015planning,de2017negotiating} model the nested beliefs (\eg, ``I believe that you believe that I believe...'') and simulate the opponents' reasoning processes to predict their actions.
Unlike these existing methods, which usually require a large amount of interactive data to generate useful opponent models, our L2E framework does not explicitly model the opponent and acquires the ability to exploit different opponents by training with limited interactions with different styles of opponents.

\subsection{Meta-Learning}
\label{meta_learning}
Meta-learning is a new trend of research in the machine learning community that tackles learning to learn~\cite{hospedales2020meta}.
It leverages experiences in the training phase to learn how to learn, acquiring the ability to generalize to new environments or new tasks.
Recent progress in meta-learning has achieved impressive results ranging from classification and regression in supervised learning~\cite{finn2017model,nichol2018first} to new task adaption in reinforcement learning~\cite{wang2016learning,xu2018meta}.
Some recent works have also initially explored the application of meta-learning in opponent modeling.
For example, the theory of mind network (ToMnet)~\cite{rabinowitz2018machine} uses meta-learning to improve the predictions about the opponents' future behaviors.
Another related work~\cite{al2017continuous} uses meta-learning to handle the non-stationarity problem in multi-agent interactions.
Unlike these methods, we focus on improving the agents' ability to quickly adapt to different unknown opponents.

\subsection{Strategy Generation}
\label{SG}
The automatic generation of effective opponent strategies for training is a critical step in our approach. Furthermore, how to generate diverse strategies has been preliminarily studied in the reinforcement learning community. In specific, diverse strategies can be obtained in various ways, including adding some diversity regularization to the optimization objective~\cite{abdullah2019wasserstein}, randomly searching in some diverse parameter space~\cite{plappert2017parameter,fortunato2017noisy}, using information-based strategy proposal~\cite{eysenbach2018diversity,gupta2018unsupervised}, and searching diverse strategies with evolutionary algorithms~\cite{agapitos2008generating,wang2019paired,jaderberg2017population,jaderberg2019human}.
More recently, researchers from DeepMind propose a league training paradigm to obtain a Grandmaster level StarCraft II AI (\ie, AlphaStar) by training a diverse league of continually adapting strategies and counter-strategies~\cite{vinyals2019grandmaster}.
Different from AlphaStar, our opponent strategy generation algorithm exploits adversarial training and diversity-regularized policy optimization to produce challenging and diverse opponents, respectively.

\section{Method}
This paper proposes a novel L2E framework to endow the agents to adapt to diverse opponents quickly.
As shown in Fig.~\ref{overview}, L2E mainly consists of two modules, \ie, the base policy training part and the opponent strategy generation part.
In the base policy training part, our goal is to find a base policy that, given the unknown opponent, can fast adapt to it by using only a few interactions.
To this end, the base policy is trained to be able to adapt to diverse opponents.
In specific, giving an opponent $O$, the base policy $B$ first adjusts itself to obtain an adapted policy $B'$ by using a little interaction data between $O$ and $B$, the base policy is then optimized to maximize the rewards that $B'$ gets when facing $O$. In other words, the base policy has learned how to adapt to its opponents and exploit them quickly.

The opponent strategy generation provides the base policy training part with challenging and diverse training opponents automatically.
First, our proposed opponent strategy generation (OSG) algorithm can produce difficult-to-exploit opponents.
In specific, the base policy $B$ first adjusts itself to obtain an adapted policy $B'$ using a little interaction data between $O$ and $B$, the opponent $O$ is then optimized to \emph{minimize} the rewards that $B'$ gets when facing $O$.
The resulting opponent $O$ is hard to exploit since even if the base policy $B$ adapts to $O$, the adapted policy $B'$ cannot take advantage of $O$.
By training with these hard to exploit opponents, the base policy can eliminate the weakness in its adaptability and improve its robustness effectively.
Second, our OSG algorithm can further produce diverse training opponents with a novel diversity-regularized policy optimization procedure.
More specifically,
we first formalize the difference between opponent policies as the difference between the distribution of trajectories induced by each policy.
The difference between distributions can be evaluated by the Maximum Mean Discrepancy (MMD) metric~\cite{gretton2007kernel}.
Then, MMD is integrated as a regularization term in the policy optimization process to identify various opponent policies.
By training with these diverse opponents, the base policy's generalization ability is improved significantly.
Next, we introduce these two modules in detail.

\subsection{Base Policy Training}
\label{Base Policy Training}

Our goal is to find a base policy $B$ that can fast adapt to an unknown opponent $O$ by updating the parameters of $B$ using only a few interactions between $B$ and $O$.
The key idea is to train the base policy $B$ against many opponents to maximize its payoffs by using only a small amount of interactive data during training, such that it acquires the ability to exploit different opponents quickly.
In effect, our L2E framework treats each opponent as a training example.
After training, the resulting base policy $B$ can quickly adapt to new and unknown opponents using only a few interactions.
Without loss of generality, the base policy $B$ is modeled by a deep neural network in this work, \ie, a parameterized function $\pi_\theta$ with parameters $\theta$.
Similarly, the opponent $O$ for training is also a deep neural network $\pi_\phi$ with parameters $\phi$.
We model the base policy as playing against an opponent in a two-player Markov game~\cite{shapley1953stochastic}.
This Markov game $M = (S, (A_B,A_O),T,(R_B,R_O))$ consists of the state space $S$, the action space $A_B$ and $A_O$, and a state transition function $T: S \times A_B \times A_O \to \Delta\left( S \right)$ where $\Delta\left( S \right)$ is a probability distribution on $S$.
The reward function $R_i: S \times A_B \times A_O \times S \to \mathbb{R} $ for each player $i \in \{B, O\} $ depends on the current state, the next state and both players' actions.
Given a training opponent $O$ whose policy is known and fixed, this two-player Markov game $M$ reduces to a single-player Markov Decision Process (MDP), \ie, $M_B^O = (S, A_B, T_B^O, R_B^O)$.
The state and action space of $M_B^O$ are the same as in $M$.
The transition and reward functions have the opponent policy embedded:
\begin{equation}
T_B^O(s, a_B) = T(s, a_B, a_O)
\label{transition}
\end{equation}
\begin{equation}
R_B^O(s,a_B,s') = R_B(s,a_B, a_O, s')
\label{reward}
\end{equation}
where the opponent's action is sampled from its policy $a_O \sim \pi_\phi(\cdot \mid s ) $\footnote{For brevity, we assume that environment is fully observable. More generally, agents use their observations to make decisions.}.
Throughout the paper, $M_X^Y$ represents a single-player MDP, which is reduced from a two-player Markov game (\ie, player $X$ and player $Y$). In this MDP, the player $Y$ is fixed and can be regarded as part of the environment.

Suppose a set of training opponents $\{O_i\}_{i=1}^N$ is given. For each training opponent $O_i$, an MDP $M_B^{O_i}$ can be constructed as described above.
The base policy $B$, \ie, $\pi_\theta$ is allowed to query a limited number of sample trajectories $\tau$ to adapt to $O_i$.
In our method, the adapted parameters $\theta^{O_i}$ of the base policy are computed using one or more gradient descent updates with the sample trajectories $\tau$. For example, when using one gradient update:
\begin{equation}
\theta^{O_i} = \theta - \alpha \nabla_\theta \mathcal{L}_B^{O_i}(\pi_\theta),
\label{one_step_adapt_part1}
\end{equation}
\begin{equation}
\mathcal{L}_B^{O_i}(\pi_\theta) = - \mathbb{E}_{\tau \sim M_B^{O_i}}[\sum\nolimits_t \gamma^t R_B^{O_i}(s^{(t)},a_B^{(t)}, s^{(t+1)})].
\label{one_step_adapt_part2}
\end{equation}
$\tau \sim M_B^{O_i}$ represents that the trajectory $\tau = \{ s^{(1)},a_B^{(1)},s^{(2)}, \ldots, s^{(t)},a_B^{(t)},s^{(t+1)}, \ldots \} $ is sampled from the MDP $M_B^{O_i}$, where $ s^{(t+1)} \sim T_B^{O_i}(s^{(t)}, a_B^{(t)}) $ and $a_B^{(t)} \sim \pi_\theta(\cdot \mid s^{(t)} ) $.

We use $B^{O_i}$ to denote the updated base policy, \ie, $\pi_{\theta^{O_i}}$.
$B^{O_i}$ can be seen as an \emph{opponent-specific} policy, which is updated from the base policy through fast adaptation.
Our goal is to find a generalizable base policy whose opponent-specific policy $B^{O_i}$ can exploit its opponent $O_i$ as much as possible.
To this end, we optimize the parameters $\theta$ of the base policy to maximize the rewards that $B^{O_i}$ gets when interacting with $O_i$.
More concretely, the \emph{learning to exploit} objective function is defined as follows:
\begin{align}
\begin{aligned}
&\min_\theta \sum\nolimits_{i=1}^{N} { \mathcal{L}_{B^{O_i}}^{O_i}(\pi_{\theta^{O_i}}) } \\
&= \min_\theta \sum\nolimits_{i=1}^{N} { \mathcal{L}_{B^{O_i}}^{O_i}( \pi_{ \theta - \alpha \nabla_\theta \mathcal{L}_B^{O_i}(\pi_\theta) } } ).
\end{aligned}
\end{align}
It is worth noting that the optimization is performed over the base policy's parameters $\theta$, whereas the objective is computed using the adapted based policy's parameters $\theta^{O_i}$. The parameters $\theta$ of the base policy are updated as follows:
\begin{equation}
\theta = \theta - \beta \nabla_\theta \sum\nolimits_{i=1}^{N} { \mathcal{L}_{B^{O_i}}^{O_i}(\pi_{\theta^{O_i}}) }.
\label{one_step_adapt_part3}
\end{equation}
In effect, our L2E framework aims to find a base policy that can significantly exploit the opponent with only a few interactions with it (\ie, with a few gradient steps).
The resulting base policy has learned how to adapt to different opponents and exploit them quickly.
An overall description of the base policy training procedure is shown in Alg.~\ref{alg:base_policy_train} which consists of three main steps.
First, generating hard-to-exploit opponents through the Hard-OSG module (Section~\ref{Hard-to-Exploit Opponents Generation}).
Second, generating diverse opponent policies through the Diverse-OSG module (Section~\ref{Diverse Opponents Generation}).
Third, training the base policy with these opponents to obtain fast adaptability.

\begin{algorithm}[htb]
	\caption{L2E's base policy training procedure.}
	\label{alg:base_policy_train}
	\begin{algorithmic}
		\STATE {\bfseries Input:} Step size hyper parameters $\alpha$, $\beta$; base policy $B$ with parameters $\theta$; opponent policy $O$ with parameters $\phi$.
		\STATE {\bfseries Output:} An adaptive base policy $B$.
		\STATE Randomly initialize $\theta$, $\phi$;
		\STATE Initialize policy pool $ \mathcal{M}=\{O\}$;
		\FOR{$1 \le e \le epochs$}
			\STATE $O$ = \textbf{Hard-OSG($B$)}; \qquad \qquad \quad $\triangleright$(Section~\ref{Hard-to-Exploit Opponents Generation})
			\STATE $\mathcal{P} = $\textbf{Diverse-OSG($B,O,N$)}; \qquad $\triangleright$(Section~\ref{Diverse Opponents Generation})
			\STATE Update opponent pool $\mathcal{M}= \mathcal{M} \cup \mathcal{P}$;
			\STATE Sample a batch of opponents $O_i \sim \mathcal{M}$;
			\FOR{Each opponent $ O_i$}
				\STATE Construct a single-player MDP $M_B^{O_i}$;
				\STATE Use Eq.~(\ref{one_step_adapt_part1}) to obtain an adapted policy $B^{O_i}$;
				\ENDFOR \\
			Update $B$'s parameters $\theta$ according to Eq.~(\ref{one_step_adapt_part3}).
	\ENDFOR
	\end{algorithmic}
\end{algorithm}

\subsection{Automatic Opponent Generation}
Previously, we assumed that the set of opponents had been given.
How to automatically generate effective opponents for training is the key to the success of our L2E framework.
The training opponents should be challenging enough (\ie, hard to exploit). By learning to exploit these hard-to-exploit opponents, the base policy $B$ can eliminate the weakness in its adaptability and become more robust.
Besides, they should be sufficiently diverse. The more diverse they are, the stronger the generalization ability of the resulting base policy.
We propose a novel opponent strategy generation (OSG) algorithm to achieve these goals.

\subsubsection{Hard-to-Exploit Opponents Generation}
\label{Hard-to-Exploit Opponents Generation}

We use the idea of adversarial learning to generate challenging training opponents for the base policy $B$.
From the perspective of the base policy $B$, giving an opponent $O$, $B$ first adjusts itself to obtain an adapted policy, \ie, the opponent-specific policy $B^{O}$, the base policy is then optimized to \emph{maximize} the rewards that $B^{O}$ gets when interacting with $O$.
Contrary to the base policy's goal, we want to find a hard-to-exploit opponent $\widehat{O}$ for the current base policy $B$, such that even if $B$ adapts to $\widehat{O}$, the adapted policy $B^{\widehat{O}}$ cannot take advantage of $\widehat{O}$. In other words, the hard-to-exploit opponent $\widehat{O}$ is trained to \emph{minimize} the rewards that $B^{\widehat{O}}$ gets when interacting with $\widehat{O}$.
The base policy attempts to increase its adaptability by learning to exploit different opponents, while the hard-to-exploit opponent adversarially tries to minimize the base policy's adaptability, \ie, maximize its \emph{counter-adaptability}.

More concretely, the hard-to-exploit opponent $\widehat{O}$ is also a deep neural network $\pi_{\widehat{\phi}}$ with randomly initialized parameters $\widehat{\phi}$.
At each training iteration, an MDP $M_B^{\widehat{O}}$ can be constructed.
The base policy $B$ first query a limited number of trajectories to adapt to $\widehat{O}$.
The parameters $\theta^{\widehat{O}}$ of the adapted policy $B^{\widehat{O}}$ are computed using one gradient descent update,
\begin{equation}
\theta^{\widehat{O}} = \theta - \alpha \nabla_\theta \mathcal{L}_B^{\widehat{O}}(\pi_\theta).
\label{hard_one_step_adapt_part1}
\end{equation}
\begin{equation}
\mathcal{L}_B^{\widehat{O}}(\pi_\theta) = - \mathbb{E}_{\tau \sim M_B^{\widehat{O}}}[\sum\nolimits_t \gamma^t R_B^{\widehat{O}}(s^{(t)},a_B^{(t)}, s^{(t+1)})].
\label{hard_one_step_adapt_part2}
\end{equation}
$\widehat{O}$'s parameters $\widehat{\phi}$ are optimized to minimize the rewards that $B^{\widehat{O}}$ gets when interacting with $\widehat{O}$. This is equivalent to maximizing the rewards that $\widehat{O}$ gets since we consider the two-player zero-sum competitive setting in this work.
More concretely, the parameters $\widehat{\phi}$ are updated as follows:
\begin{equation}
\widehat{\phi} = \widehat{\phi} - \alpha \nabla_{\widehat{\phi}} { \mathcal{L}_{\widehat{O}}^{B^{\widehat{O}}}(\pi_{\widehat{\phi}}) }
\label{hard_step2_part1}
\end{equation}
\begin{equation}
\mathcal{L}_{\widehat{O}}^{B^{\widehat{O}}}(\pi_{\widehat{\phi}}) = - \mathbb{E}_{\tau' \sim M_{\widehat{O}}^{B^{\widehat{O}}}} [\sum\nolimits_t \gamma^t R_{\widehat{O}}^{B^{\widehat{O}}}(s^{(t)},a_{\widehat{O}}^{(t)}, s^{(t+1)})].
\label{hard_step2_part2}
\end{equation}
After several rounds of iteration, we can obtain a hard-to-exploit opponent $\pi_{\widehat{\phi}}$ for the current base policy $B$.
The pseudo code of this hard-to-exploit training opponent generation algorithm is shown in Appendix B.1.

\subsubsection{Diverse Opponents Generation}
\label{Diverse Opponents Generation}
Training an effective base policy requires not only the hard-to-exploit opponents but also diverse opponents of different styles.
The more diverse the opponents used for training, the stronger the generalization ability of the resulting base policy.
From a human player's perspective, the opponent style is usually defined as different types, such as aggressive, defensive, elusive, \etc.
The most significant difference between opponents with different styles lies in the actions taken in the same state.
Take poker as an example, opponents of different styles tend to take different actions when holding the same hand.
Based on the above analysis, we formalize the difference between opponent policies as the difference between the distribution of trajectories induced by each policy when interacting with the base policy.
We argue that differences in trajectories better capture the differences between different opponent policies.

Formally, given a base policy $B$, \ie, $\pi_\theta$ and an opponent policy $O_i$, \ie, $\pi_{\phi_i}$, our diversity-regularized policy optimization algorithm is to generate a new opponent $O_j$, \ie, $\pi_{\phi_j}$ whose style is different from $O_i$.
We first construct two MDPs, \ie, $M_{O_i}^{B}$ and $M_{O_j}^{B}$, and then sample two sets of trajectories, \ie, $\mathrm{T}_i = \{ \tau \sim M_{O_i}^{B} \}$ and $\mathrm{T}_j= \{ \tau \sim M_{O_j}^{B} \}$ from this two MDPs.
The stochasticity in the MDP and the policy will induce a distribution over trajectories.
We use the Maximum Mean Discrepancy (MMD) metric~\cite{gretton2007kernel} (see Appendix A for details) to measure the differences between $\mathrm{T}_i$ and $\mathrm{T}_j$:
\begin{align}\begin{aligned}
&\text{MMD}^{2}(\mathrm{T}_i,\mathrm{T}_j) = \mathbb{E}_{\tau, \tau' \sim M_{O_i}^{B} } k\left(\tau, \tau^{\prime}\right) \\
&-2 \mathbb{E}_{\tau \sim M_{O_i}^{B}, \tau' \sim M_{O_j}^{B} } k(\tau, \tau')  +\mathbb{E}_{\tau, \tau' \sim M_{O_j}^{B}} k\left(\tau, \tau^{\prime}\right).
\end{aligned}\end{align}
$k$ is the Gaussian radial basis function kernel defined over a pair of trajectories:
\begin{equation}
k(\tau,\tau') = \exp(-\frac{ \| g(\tau)- g(\tau') \|^2}{2h}).
\end{equation}
In our experiments, we found that simply setting the bandwidth $h$ to 1 produced satisfactory results.
$g$ stacks the states and actions of a trajectory into a vector.
For trajectories with different length, we clip the trajectories when both of them are longer than the minimum length $L$.
Usually, the trajectory's length does not exceed $L$, and we apply the masking based on the done signal from the environment to make them the same length.

There overall objective function of our proposed diversity-regularized policy optimization algorithm is as follows:
\begin{align}\begin{aligned}
\mathcal{L}^{\phi_i}{(\phi_j)} &= -\mathbb{E}_{\tau \sim M_{O_j}^{B}}[\sum\nolimits_t \gamma^t R_{O_j}^{B}(s^{(t)},a_{O_j}^{(t)}, s^{(t+1)})] \\
&-\alpha_\text{mmd}\operatorname{MMD}^{2}(\mathrm{T}_i,\mathrm{T}_j).
\end{aligned}\end{align}
The first term is to maximize the rewards that $O_j$ gets when interacting with the base policy $B$.
The second one measures the difference between $O_j$ and the existing opponent $O_i$.
By this diversity-regularized policy optimization, the resulting opponent $O_j$
is not only good in performance but also diverse relative to the existing policy.
Appendix A details the gradient calculation of the MMD term.

We can iteratively apply the above algorithm to find a set of $N$ distinct and diverse opponents.
In specific, subsequent opponents are learned by encouraging diversity with respect to previously generated opponent set $S$.
The distance between an opponent $O_m$ and an opponent set $S$ is defined by the distance between
$O_m$ and $O_n$, where $O_n \in S$ is the most similar policy to $O_m$.
Suppose we have obtained a set of opponents $S = \{ O_m\}_{m=1}^{M}, M<N$.
The $M\!\!+\!\!1\text{-th}$ opponent, \ie, $\pi_{\phi_{M+1}}$ can be obtained by optimizing:
\begin{align}\begin{aligned}
\label{divese_opponent}
&\mathcal{L}^{S}(\phi_{M+1}) = - \alpha_\text{mmd}  \min_{O_i \in S} {\operatorname{MMD}^{2}(\mathrm{T}_{i},\mathrm{T}_{M+1})} \\
&-\mathbb{E}_{\tau \sim M_{O_{M+1}}^{B}}[\sum\nolimits_t \gamma^t R_{O_{M+1}}^{B}(s^{(t)},a_{O_{M+1}}^{(t)}, s^{(t+1)})].
\end{aligned}\end{align}
By doing so, the resulting $M\!+\!1\text{-th}$ opponent remains diverse relative to the opponent set $S$.
Appendix B.2 provides the pseudo code of the diverse training opponent generation algorithm.

\section{Experiments}

\begin{figure}[htbp]
	\centering 
	\includegraphics[width=1.0\linewidth]{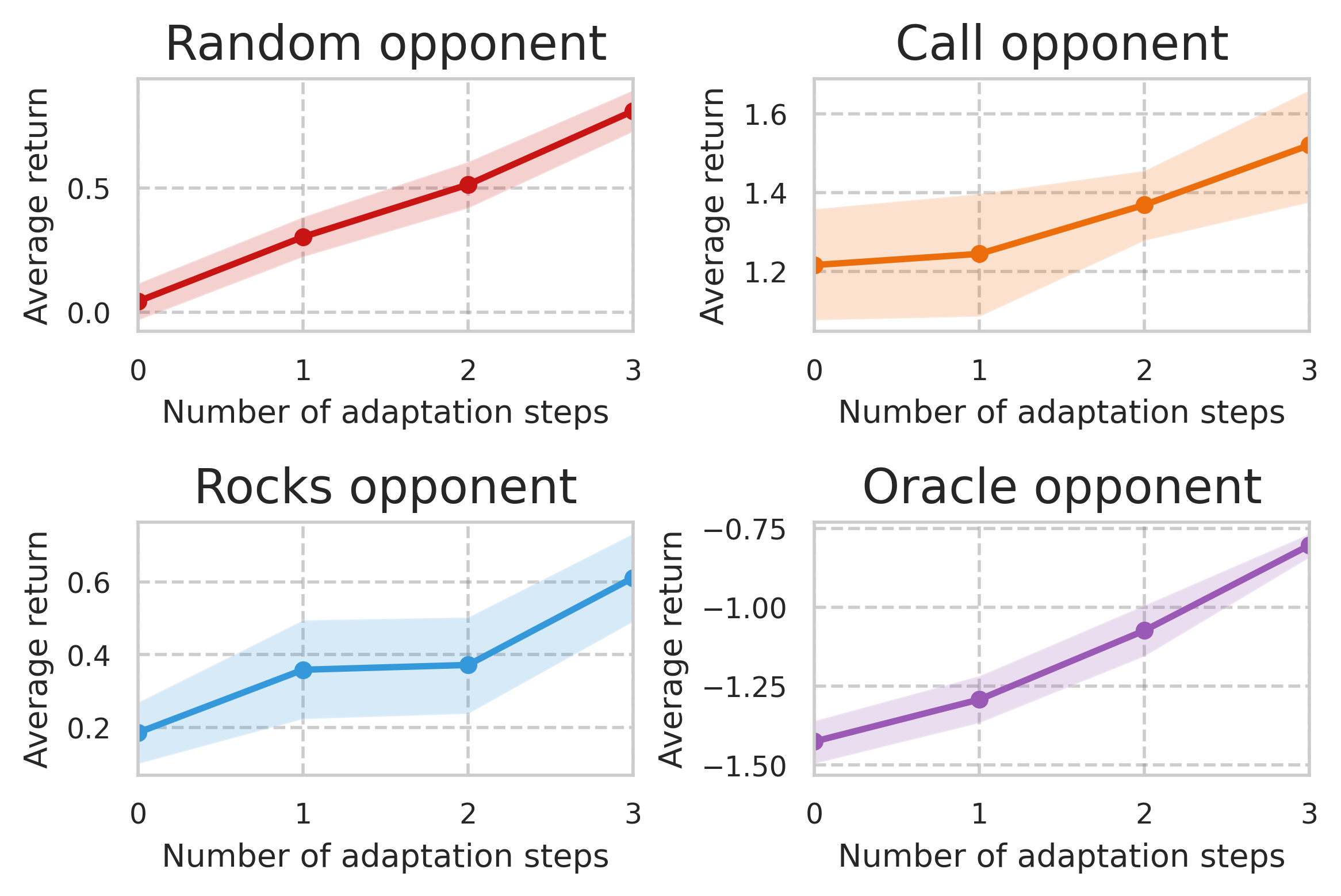}
	\caption{The trained base policy using our L2E framework can quickly adapt to different opponents of different styles and strengths in the Leduc poker environment.
	}
	\label{fig_fast}
	\vspace{-3mm}
\end{figure}

In this section, we conduct extensive experiments to evaluate the proposed L2E framework.
We evaluate algorithm performance on the Leduc poker, the BigLeduc poker, and a Grid Soccer environment, the commonly used benchmark for opponent modeling~\cite{lanctot2017unified,steinberger2019single,he2016opponent}.
We first verify that the trained base policy using our L2E framework quickly exploit a wide range of opponents with only a few gradient updates.
Then, we compare with other baseline methods to show the superiority of our L2E framework.
Finally, we conduct a series of ablation experiments to demonstrate each part of our L2E framework's effectiveness.

\subsection{Rapid Adaptability}

In this section, we verify the trained base policy's ability to quickly adapt to different opponents in the Leduc poker environment (see Appendix C for more details).
We provide four opponents with different styles and strengths.
1) The \textbf{random opponent} randomly takes actions whose strategy is relatively weak but hard to exploit since it does not have an evident decision-making style.
2) The \textbf{call opponent} always takes call actions and has a fixed decision-making style that is easy to exploit.
3) The \textbf{rocks opponent} takes actions based on its hand-strength, whose strategy is relatively strong.
4) The \textbf{oracle opponent} is a cheating and the strongest player who can see the other players' hands and make decisions based on this perfect information.

As shown in Fig.~\ref{fig_fast}, the base policy achieves a rapid increase in its average returns with a few gradient updates against all four opponent strategies.
For the call opponent, which has a clear and monotonous style, the base policy can significantly exploit it.
Against the random opponent with no clear style, the base policy can also exploit it quickly.
When facing the strong rocks opponent or even the strongest oracle opponent, the base policy can quickly improve its average returns.
One significant advantage of L2E is that the same base policy can exploit a wide range of opponents with different styles, demonstrating its strong generalization ability.
\begin{figure}[t]
	\centering 
	\includegraphics[width=0.85\linewidth]{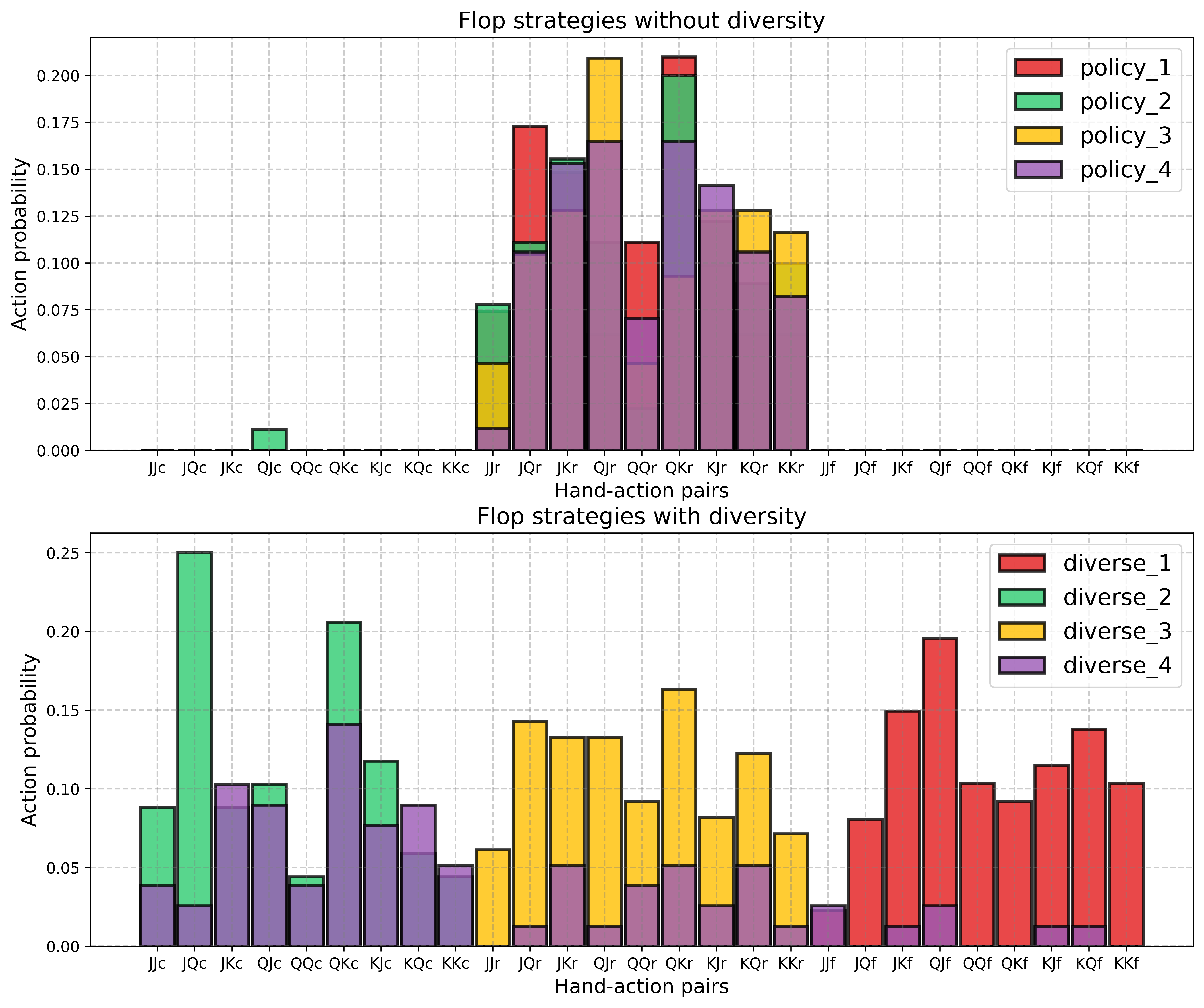} 
	\vspace{-3mm}
	\caption{Visualization of the styles of the strategies generated with or without the MMD regularization term in the Leduc poker environment. Action probability is the probability that a player will take a corresponding action with a particular hand.}
	\label{fig_visualization}
	\vspace{-3mm}
\end{figure}

\subsection{Comparisons with Other Baseline Methods}

\begin{table}[t]
	\caption{The average return of each method when performing rapid adaptation against different opponents in Leduc Poker. The adaptation process is restricted to a three-step gradient update.}
	\label{leduc avereturn}
	\centering
	\scalebox{0.7}{
		\begin{tabular}{cccccc}
			\toprule
			& Random& Call& Rocks& Nash& Oracle\\
			\midrule
			L2E& 0.42$\pm$0.32& \textbf{1.34$\pm$0.14}& \textbf{0.38$\pm$0.17}& \textbf{-0.03$\pm$0.14}& \textbf{-1.15$\pm$0.27} \\
			MAML& \textbf{1.27$\pm$0.17}& -0.23$\pm$0.22& -1.42$\pm$0.07& -0.77$\pm$0.23& -2.93$\pm$0.17 \\
			Random& -0.02$\pm$3.77& -0.02$\pm$3.31& -0.68$\pm$3.75& -0.74$\pm$4.26& -1.90$\pm$4.78 \\
			TRPO& 0.07$\pm$0.08& -0.22$\pm$0.09& -0.77$\pm$0.12& -0.42$\pm$0.07& -1.96$\pm$0.46 \\
			TRPO-P&  0.15$\pm$0.17&  -0.05$\pm$0.14&  -0.70$\pm$0.27&  -0.61$\pm$0.32&  -1.32$\pm$0.27 \\
			EOM&  0.30$\pm$0.15&  -0.01$\pm$0.05&  -0.13$\pm$0.20&  -0.36$\pm$0.11&  -1.82$\pm$0.28 \\
			\bottomrule
	\end{tabular}}
\vspace{-3mm}
\end{table}

As discussed in Section~\ref{intro}, most previous opponent modeling methods require constructing explicit opponent models from a large amount of data before learning to adapt to new opponents.
To the best of our knowledge, our L2E framework is the first attempt to use meta-learning to learn to exploit opponents without building explicit opponent models.
To demonstrate the effectiveness of L2E, we design several competitive baseline methods.
As with the previous experiments, we also use three gradient updates when adapting to a new opponent.
1) \textbf{MAML}. The seminal meta-learning algorithm MAML~\cite{finn2017model} is designed for single-agent environments.
We have redesigned and reimplemented the MAML algorithm for the two-player competitive environments.
The MAML baseline trains a base policy by continually sampling the opponent's strategies, either manually specified or randomly generated.
2) \textbf{Random}. The Random baseline is neither pre-trained nor updated online. It acts as a sanity check.
3) \textbf{TRPO}. The TRPO baseline does not perform pre-training and uses the TRPO algorithm~\cite{schulman2015trust} to update its parameters via three-step gradient updates to adapt to different opponents.
4) \textbf{TRPO-P}. The TRPO-P baseline is pre-trained by continually sampling the opponents' strategies (similar to the MAML baseline) and then fine-tuned against a new opponent.
5) \textbf{EOM}. The Explicit Opponent Modeling (EOM) baseline collects interaction data during the adaptation process to explicitly fit an opponent model $M_O$. The best response trained for $M_O$ is used to interact with the opponent again to evaluate EOM's performance.

To evaluate different algorithms more comprehensively, we additionally add a new Nash opponent.
This opponent's policy is a part of an approximate Nash Equilibrium generated iteratively by the CFR~\cite{zinkevich2008regret} algorithm.
Playing a strategy from a Nash Equilibrium in a two-player zero-sum game is guaranteed not to lose in expectation even if the opponent uses the best response strategy when the value of the game is zero.
We show the performance of the various algorithms in Table~\ref{leduc avereturn}.
L2E maintains the highest profitability against all four types of opponents other than the random type.
L2E can exploit the opponent with evident style significantly, such as the Call opponent.
Compared to other baseline methods, L2E achieved the highest average return against opponents with unclear styles, such as the Rocks opponent, the Nash opponent, and the cheating Oracle opponent.

\subsection{Ablation Studies}
\subsubsection{Effects of the Diversity-regularized Policy Optimization}

In this section, we verify whether our proposed diversity-regularized policy optimization algorithm can effectively generate policies with different styles.
In Leduc poker, hand-action pairs represent different combinations of hands and actions.
In the pre-flop phase, each player's hand has three possibilities, \ie, J, Q, and K.
Meanwhile, each player also has three optional actions, \ie, Call (c), Rise (r), and Fold (f).
For example, `Jc' means to call when getting the jack.
Fig.~\ref{fig_visualization} shows that without the MMD regularization term, the generated strategies have similar styles.
By optimizing with the MMD term, the generated strategies are diverse enough which cover a wide range of different states and actions.

\subsubsection{Effects of the Hard-OSG and the Diverse-OSG}

\begin{figure}[h] 
	\centering 
	\includegraphics[width=0.7\linewidth]{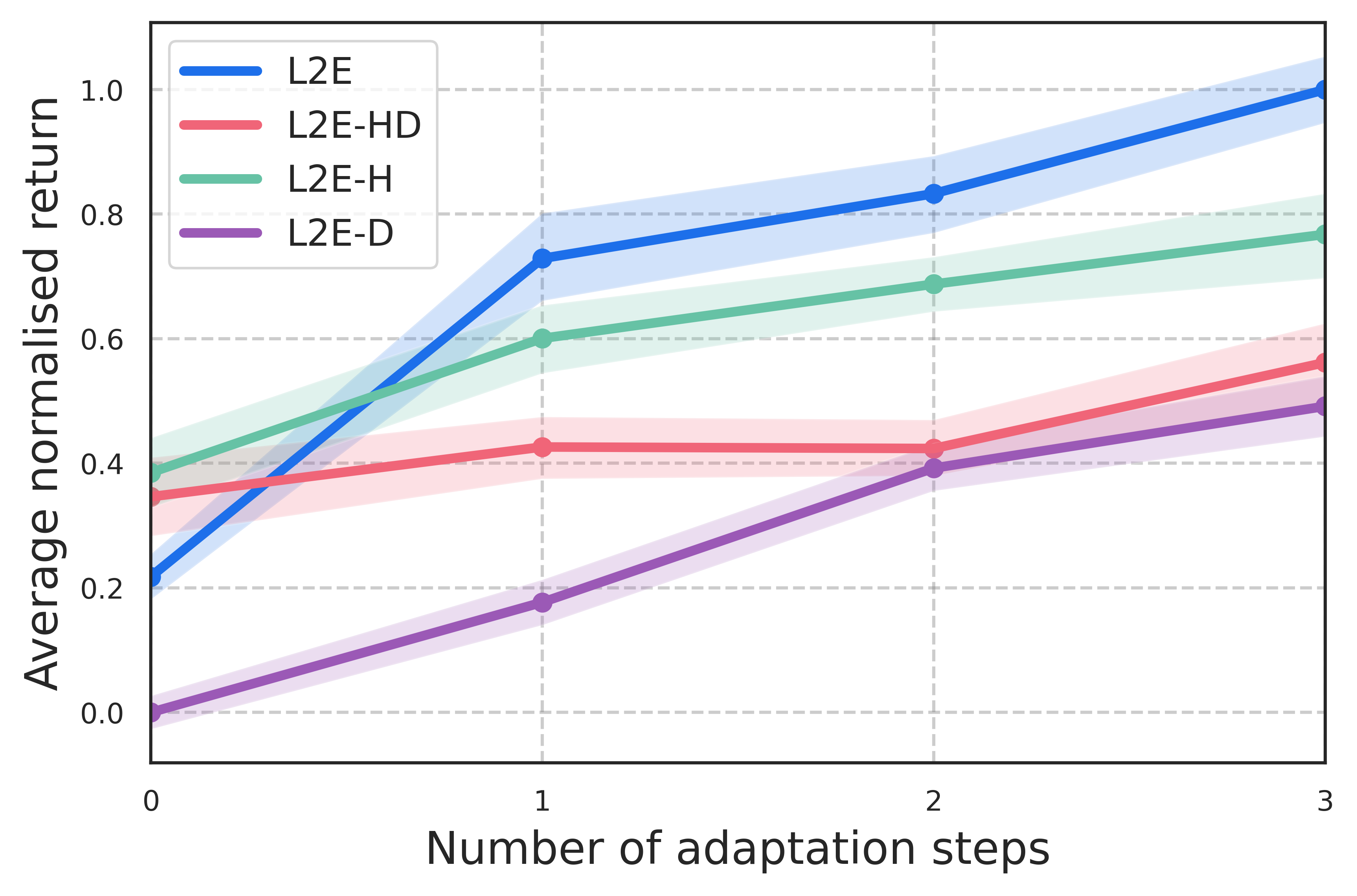}
	\vspace{-3mm}
	\caption{
		We normalized the sum of the base policy's returns when facing aggressive and defensive opponents in the soccer environment and mapped them to the range [0, 1].
		Each curve shows the average normalized returns of the base policy trained with different variants of L2E.} 
	\label{ablation fig} 
	\vspace{-5mm}
\end{figure}

As discussed previously, a crucial step in L2E is the automatic generation of training opponents.
The Hard-OSG and Diverse-OSG modules are used to generate opponents that are difficult to exploit and diverse in styles.
Fig.~\ref{ablation fig} shows each module's impact on the performance, `L2E-H' is L2E without the Hard-OSG module, `L2E-D' is L2E without the Diverse-OSG module, and `L2E-HD' removes both modules altogether.
The results show that both Hard-OSG and Diverse-OSG have a crucial influence on L2E's performance.
It is clear that the Hard-OSG module helps to enhance the stability of the base policy, and the
Diverse-OSG module can improve the base policy's performance significantly.
Specifically, the performance of `L2E-HD' is unstable, \eg, its two-step adaptation performance is roughly the same as its one-step adaptation performance.
`L2E-H' alleviates this problem, but its final performance is still worse than L2E, and the improvement eventually reaches a plateau.
With the addition of Hard-OSG, L2E achieves the best performance, and the improvement is faster and more stable.
To further demonstrate the generalization ability of L2E, we conducted a series of additional experiments on the BigLeduc poker and the Grid Soccer game environment (see Appendix D and E for more details).

\subsection{Convergence Analysis and the Relationship with Nash Equilibrium}

\begin{figure}[htbp] 
	\centering 
	\includegraphics[width=0.7\linewidth]{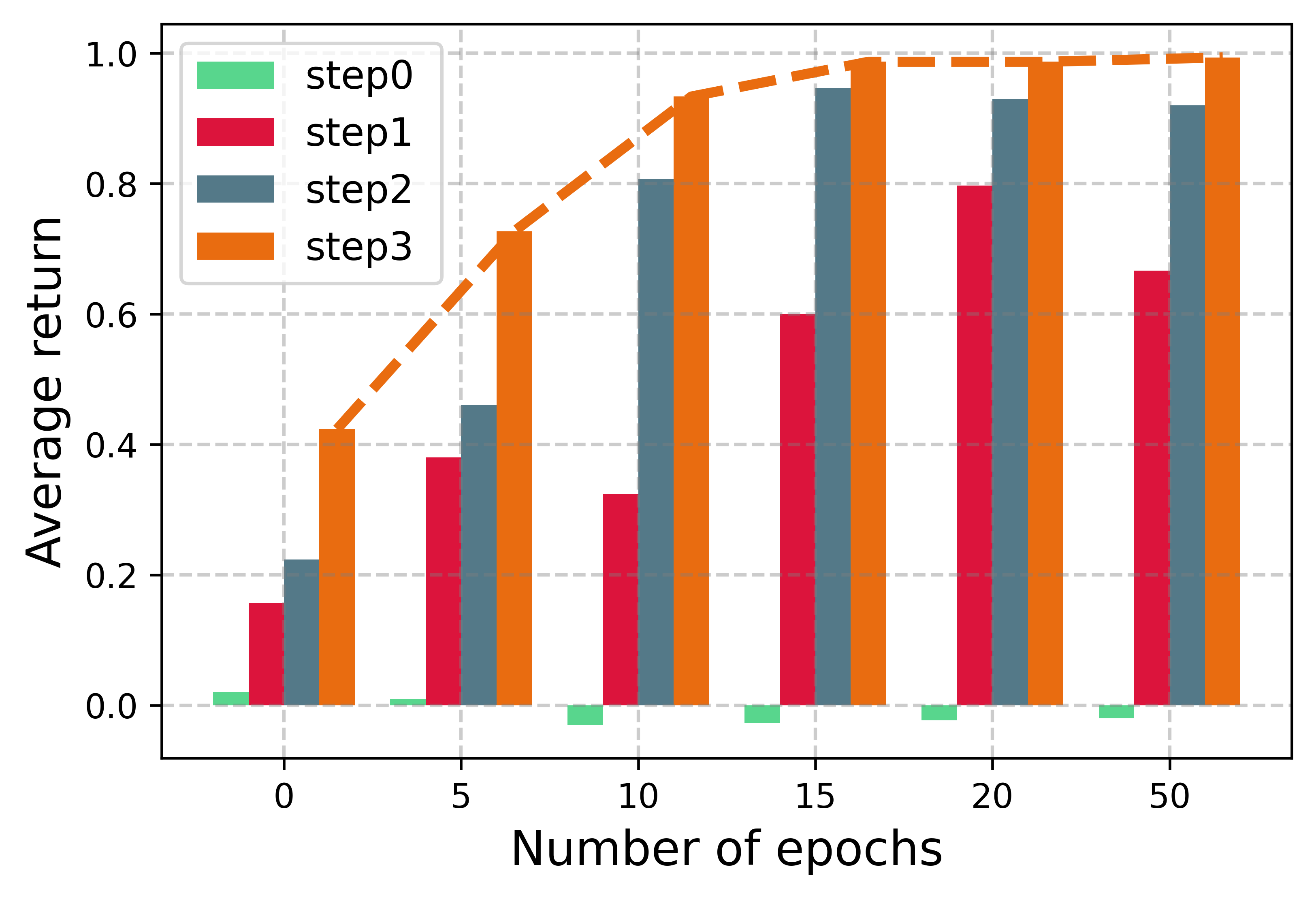}
	\vspace{-3mm}
	\caption{L2E's performance converges after a certain number of iterations.}
	\label{conv}
\end{figure}

One may wonder whether L2E is supposed to converge at all.
If L2E can converge, what is the relationship between the obtained base policy and the equilibrium concepts, such as Nash equilibrium?
Since convergence is difficult to analyze theoretically in game theory, we have designed a series of small-scale experiments to empirically analyze L2E's convergence properties using the Rock-Paper-Scissors (RPS) game.
There are several reasons why RPS is chosen:
1) RPS's strategies are easy to visualize and analyze thanks to the small state and action spaces.
2) RPS's Nash equilibrium\footnote{Choosing Rock, Paper or Scissors with equal probability.} is obvious and unique.
3) RPS is often used in game theory for theoretical analysis.
From the experimental results in Fig.~\ref{conv}, it can be seen that as the training progresses, L2E's adaptability becomes stronger and stronger. After reaching a certain number of iterations, the improvement eventually reaches a plateau, which provides some empirical evidence for the convergence of L2E.
Similar performance convergence curves can also be observed in the Leduc poker, the BigLeduc poker, and the Grid Soccer environments.

\begin{figure}[htbp] 
	\centering 
	\includegraphics[width=0.75\linewidth]{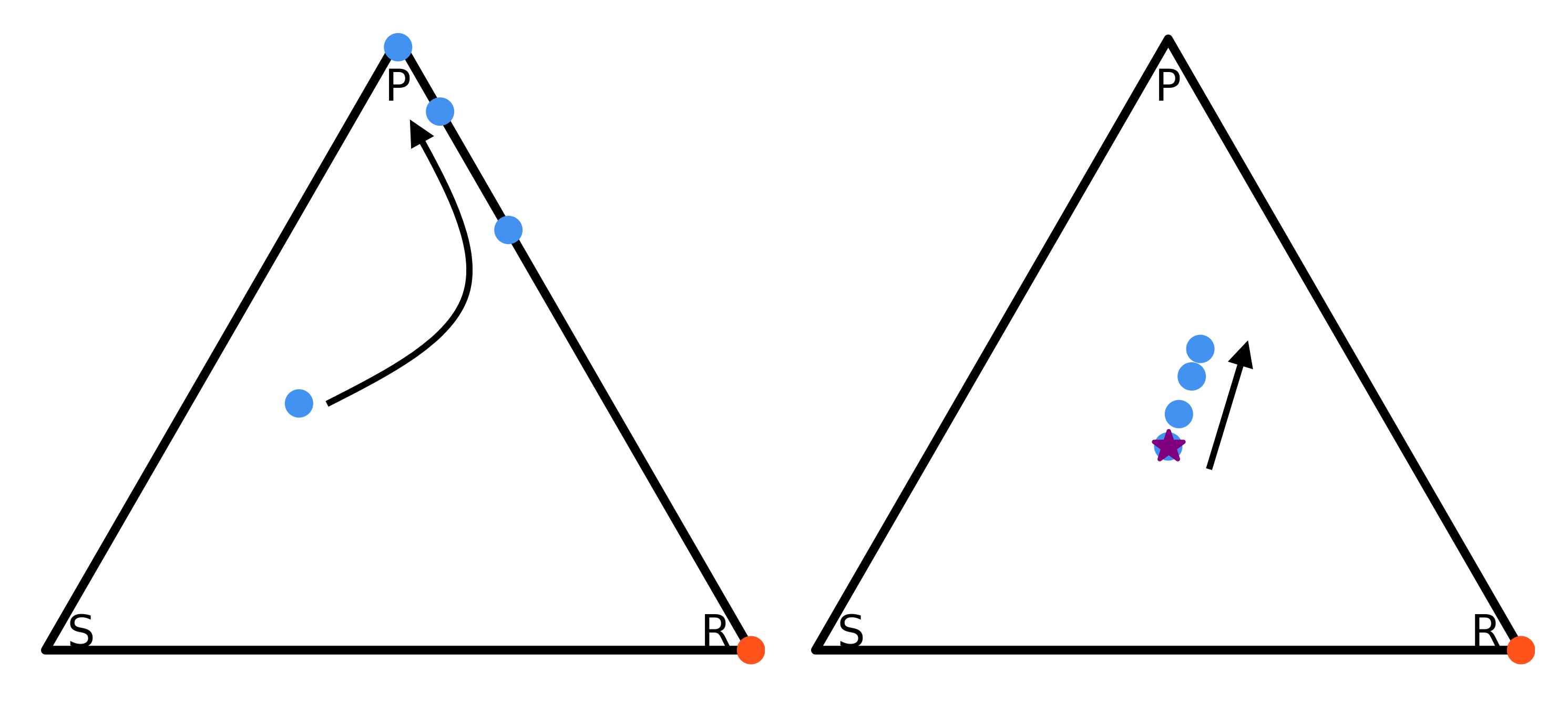}
	\vspace{-3mm}
	\caption{
		The Relationship between L2E and Nash Equilibrium.
	}
	\label{abcd} 
\end{figure}

The left side of Fig.~\ref{abcd} visualizes L2E's trained base policy, \ie, the blue point in the triangle.
It is clear that the base policy does not converge to the Nash equilibrium (\ie, the exact center of the triangle) but converges to its vicinity.
The orange dot represents the opponent who always chooses Rock.
The base policy can exploit this opponent with three steps of gradient update and quickly converge to the best response strategy (\ie, always choose Paper).
The right side of Fig.~\ref{abcd} shows that if we set the base policy to the Nash strategy by imitation learning, this \emph{Nash base policy} (\ie, the star in the triangle) is difficult to adapt to its opponent quickly.
From these results, it is clear that L2E's base policy can adapt to the opponents more quickly than the Nash base policy.
This phenomenon is easy to explain.
L2E's base policy is explicitly trained such that a few interactions with a new opponent will produce an opponent-specific policy to exploit this opponent effectively.
In contrast, the Nash base policy is essentially a random strategy in this example. Therefore, this vanilla random policy without explicitly trained to fast adapt to its opponents cannot do so.
Of course, the Nash base policy can finally converge to its opponent's best response strategy. But it is much slower than L2E's base policy.

\section{Conclusion}
We propose a Learning to Exploit (L2E) framework to exploit sub-optimal opponents without building explicit opponent models.
L2E acquires the ability to exploit opponents by a few interactions with different opponents during training to adapt to new opponents during testing quickly.
We propose a novel opponent strategy generation algorithm that produces challenging and diverse training opponents for L2E automatically.
Detailed experimental results in three challenging environments demonstrate the effectiveness of the proposed L2E framework.

\nocite{gretton2007kernel}
\nocite{sriperumbudur2010non}
\nocite{fukumizu2009kernel,gretton2012kernel}
\bibliography{reference}
\bibliographystyle{icml2021}

\appendix
\newpage
\onecolumn
\part*{Appendices}

\section{Maximum Mean Discrepancy}
\label{appendix_mmd}
We use the Maximum Mean Difference (MMD) \cite{gretton2007kernel} metric to measure the differences between the distributions of trajectories induced by different opponent strategies. In contrast to the Wasserstein distance and Dudley metrics, the MMD metric has closed-form solutions. Furthermore, unlike KL-divergence, the MMD metric is strongly consistent while exhibiting good rates of convergence \cite{sriperumbudur2010non}.

$\textbf{Definition 1}$ \emph{Let $\mathcal{F}$ be a function space $f: \mathcal{X} \rightarrow \mathbb{R}$. Suppose we have two distributions $p$ and $q$, $X:=\{x_1,...,x_m\} \sim p$, $Y:=\{y_1,...,y_n\} \sim q$.
		The MMD between $p$ and $q$ using test functions from the function space $\mathcal{F}$ is defined as follows:}
\begin{align}
	\operatorname{MMD}[\mathcal{F}, p, q]:=\sup _{f \in \mathcal{F}}\left(\mathbf{E}_{x \sim p}[f(x)]-\mathbf{E}_{y \sim q}[f(y)]\right).
\end{align}
By picking a suitable function space $\mathcal{F}$, we get the following important theorem \cite{gretton2007kernel}.

$\textbf{Theorem 1}$ \emph{Let $\mathcal{F} = \left\{f \mid\|f\|_{\mathcal{H}} \leq 1\right\}$ be a unit ball in a Reproducing Kernel Hilbert Space (RKHS). Then $\operatorname{MMD}[\mathcal{F}, p, q] = 0$ if and only if $p=q$.}

$\varphi$ is a feature space mapping from $x$ to RKHS, we can easily calculate the MMD distance using the kernel method $k(x,x'):= \langle\mathcal{\varphi}(x),\mathcal{\varphi}(x')\rangle_{\mathcal{H}}$:
\begin{align}\begin{aligned}
\label{appendix_mmd1}
\operatorname{MMD}^{2}(\mathcal{F},p, q)\\=&\left\|\mathbb{E}_{X \sim p} \varphi(X)-\mathbb{E}_{Y \sim q} \varphi(Y)\right\|_{\mathcal{H}}^{2} \\
=&\left\langle\mathbb{E}_{X \sim p} \varphi(X) - \mathbb{E}_{Y \sim q} \varphi(Y) , \mathbb{E}_{X \sim p} \varphi(X) - \mathbb{E}_{Y \sim q} \varphi(Y) \right\rangle \\
=&\mathbb{E}_{X, X^{\prime} \sim p} k\left(X, X^{\prime}\right)-2 \mathbb{E}_{X \sim p, Y \sim q} k(X, Y)+\mathbb{E}_{Y, Y^{\prime} \sim q} k\left(Y, Y^{\prime}\right).
\end{aligned}\end{align}

The expectation terms in Eq.~(\ref{appendix_mmd1}) can be approximated using samples:
\begin{align}\begin{aligned}
\begin{array}{c}
\operatorname{MMD}^{2}[\mathcal{F}, X, Y]=\frac{1}{m(m-1)} \sum_{i \neq j}^{m} k\left(x_{i}, x_{j}\right) \\ \\
+\frac{1}{n(n-1)} \sum_{i \neq j}^{n} k\left(y_{i}, y_{j}\right)-\frac{2}{m n} \sum_{i, j=1}^{m, n} k\left(x_{i}, y_{j}\right).
\end{array}
\end{aligned}
\end{align}
The gradient of the MMD term with respect to the policy's parameter 
\begin{align}
\begin{aligned}
\nabla_{\phi_j} {\mathrm{MMD}^2}(\mathrm{T}_i,\mathrm{T}_j) &=\nabla_{\phi_j} {\mathrm{MMD}^2}(\{\tau \sim M_{O_i}^{B}\},\{\tau \sim M_{O_j}^{B}\}) \\
&=\mathbb{E}_{\tau, \tau' \sim M_{O_i}^{B}}[k\left(\tau, \tau^{\prime}\right) \nabla_{\phi_j} \log (p(\tau)p(\tau^{\prime}))] \\
&-2 \mathbb{E}_{\tau \sim M_{O_i}^{B},\tau^{\prime} \sim M_{O_j}^{B}}[k\left(\tau, \tau^{\prime}\right) \nabla_{\phi_j} \log (p(\tau)p(\tau^{\prime}))] \\
&+\mathbb{E}_{\tau, \tau' \sim M_{O_j}^{B}}[k\left(\tau, \tau^{\prime}\right) \nabla_{\phi_j} \log  (p(\tau)p(\tau^{\prime}))],
\end{aligned}
\end{align}
where $p(\tau)$ is the probability of the trajectory.
Since $\mathrm{T}_i = \{\tau \sim M_{O_i}^{B}\}$, $O_i$ is the known opponent policy that has no dependence on $\phi_j$.
The gradient with respect to the parameters $\phi_j$ in first term is 0.
The gradient of the second and third terms can be easily calculated as follows:
\begin{equation}
\nabla_{\phi_j} \log(p(\tau)) = \sum\nolimits_{t=0}^{T} \nabla_{\phi_j}\log{\pi_{\phi_j}}(a_t|s_t).
\end{equation}

\section{Algorithm}

\subsection{Hard-OSG}

Alg.~\ref{alg:hard_opp_generation} is the overall description of the hard-to-exploit training opponent generation algorithm.

\begin{equation}
	\theta^{\widehat{O}} = \theta - \alpha \nabla_\theta \mathcal{L}_B^{\widehat{O}}(\pi_\theta).
	\label{append_hard_one_step_adapt_part1}
\end{equation}
\begin{equation}
	\mathcal{L}_B^{\widehat{O}}(\pi_\theta) = - \mathbb{E}_{\tau \sim M_B^{\widehat{O}}}[\sum\nolimits_t \gamma^t R_B^{\widehat{O}}(s^{(t)},a_B^{(t)}, s^{(t+1)})].
	\label{append_hard_one_step_adapt_part2}
\end{equation}
\begin{equation}
	\widehat{\phi} = \widehat{\phi} - \alpha \nabla_{\widehat{\phi}} { \mathcal{L}_{\widehat{O}}^{B^{\widehat{O}}}(\pi_{\widehat{\phi}}) }
	\label{append_hard_step2_part1}
\end{equation}
\begin{equation}
	\mathcal{L}_{\widehat{O}}^{B^{\widehat{O}}}(\pi_{\widehat{\phi}}) = - \mathbb{E}_{\tau' \sim M_{\widehat{O}}^{B^{\widehat{O}}}} [\sum\nolimits_t \gamma^t R_{\widehat{O}}^{B^{\widehat{O}}}(s^{(t)},a_{\widehat{O}}^{(t)}, s^{(t+1)})].
	\label{append_hard_step2_part2}
\end{equation}

\begin{algorithm}[htb]
	\caption{Hard-OSG, the hard-to-exploit training opponent generation algorithm.}
	\label{alg:hard_opp_generation}
	\begin{algorithmic}
		\STATE {\bfseries Input:} The latest base policy $B$ with parameters $\theta$.
		\STATE {\bfseries Output:} A hard-to-exploit opponent $\widehat{O}$ for $B$.
		\STATE Randomly initialize $\widehat{O}$'s parameters $\widehat{\phi}$;
		\FOR{$1 \le i \le epochs$}
		\STATE Construct a single-player MDP $M_B^{\widehat{O}}$;
		\STATE Sample a small number of trajectories $\tau \sim M_B^{\widehat{O}}$ using $B$ against $\widehat{O}$;
		\STATE Use Eq.~\ref{append_hard_one_step_adapt_part1} to update the parameters of $B$ to obtain an adapted policy $B^{\widehat{O}}$;
		\STATE Sample trajectories $\tau' \sim M_{\widehat{O}}^{B^{\widehat{O}}}$ using $\widehat{O}$ against $B^{\widehat{O}}$;
		\STATE Update the parameters $\widehat{\phi}$ of $\widehat{O}$ according to Eq.~\ref{append_hard_step2_part1}.
		\ENDFOR
	\end{algorithmic}
\end{algorithm}

\subsection{Diverse-OSG}

Alg.~\ref{alg:diverse_generate} is the overall description of the diverse training opponent generation algorithm.

\begin{align}\begin{aligned}
		\label{append_divese_opponent}
		&\mathcal{L}^{S}(\phi_{M+1}) = - \min_{O_i \in S} {\operatorname{MMD}^{2}(\mathrm{T}_{i},\mathrm{T}_{M+1})} \\
		&-\mathbb{E}_{\tau \sim M_{O_{M+1}}^{B}}[\sum\nolimits_t \gamma^t R_{O_{M+1}}^{B}(s^{(t)},a_{O_{M+1}}^{(t)}, s^{(t+1)})].
\end{aligned}\end{align}

\begin{algorithm}[htb]
	\caption{Diverse-OSG, the proposed diversity-regularized policy optimization algorithm to generate diverse training opponents.}
	\label{alg:diverse_generate}
	\begin{algorithmic}
		\STATE {\bfseries Input:} The latest base policy $B$, an existing opponent $O_1$, the total number of opponents that to be generated $N$.
		\STATE {\bfseries Output:}  A set of diverse opponent $S = \{ O_m\}_{m=1}^{N}$.
		\STATE Initialize the opponent set $S=\{O_1\}$;
		\FOR{ $i = 2 $ to $N$}
		\STATE Randomly initialize an opponent $O_i$'s parameters $\phi_{i}$;
		\FOR{$1 \le t \le steps$}
		\STATE Calculate the objective function $\mathcal{L}^{S}(\phi_{i})$ according to Eq.~\ref{append_divese_opponent};
		\STATE Calculate the gradient $\nabla_{\phi_{i}} \mathcal{L}^{S}(\phi_{i}) $;
		\STATE Use this gradient to update $\phi_{i}$;
		\ENDFOR
		\STATE Update the opponent set $S$, \ie, $S = S \cup O_i $.
		\ENDFOR
	\end{algorithmic}
\end{algorithm}

\subsection{L2E's testing procedure}
Alg.~\ref{alg:base_policy_test} is the testing procedure of our L2E framework.

\begin{equation}
	\theta^{O_i} = \theta - \alpha \nabla_\theta \mathcal{L}_B^{O_i}(\pi_\theta),
	\label{append_one_step_adapt_part1}
\end{equation}

\begin{algorithm}[htb]
	\caption{The testing procedure of our L2E framework.}
	\label{alg:base_policy_test}
	\begin{algorithmic}
		\STATE {\bfseries Input:} Step size hyper-parameters $\alpha$; The trained base policy $B$ with parameters $\theta$; An unknown opponent $O$.
		\STATE {\bfseries Output:} The updated base policy $B^{O}$ that has been adapted to the opponent $O$.
		\STATE Construct a single-player MDP $M_B^{O}$;
		\FOR{$1 \le  step \le steps$}
		\STATE Sample trajectories $\tau$ from the MDP $M_B^{O}$;
		\STATE Update the parameters $\theta$ of $B$ according to Eq.~\ref{append_one_step_adapt_part1}.
		\ENDFOR
	\end{algorithmic}
\end{algorithm}

\section{Leduc Poker}
\label{app_leduc_poker}
The Leduc poker generally uses a deck of six cards that includes two suites, each with three ranks (Jack, Queen, and King of Spades, Jack, Queen, and King of Hearts).
The game has a total of two rounds. Each player is dealt with a private card in the first round, with the opponent's deck information hidden.
In the second round, another card is dealt with as a community card, and the information about this card is open to both players.
If a player's private card is paired with the community card, that player wins the game; otherwise, the player with the highest private card wins the game.
Both players bet one chip into the pot before the cards are dealt.
Moreover, a betting round follows at the end of each dealing round.
The betting wheel alternates between two players, where each player can choose between the following actions: call, check, raise, or fold.
If a player chooses to call, that player will need to increase his bet until both players have the same number of chips.
If one player raises, that player must first make up the chip difference and then place an additional bet.
Check means that a player does not choose any action on the round but can only check if both players have the same chips.
If a player chooses to fold, the hand ends, and the other player wins the game.
When all players have equal chips for the round, the game moves on to the next round. The final winner wins all the chips in the game.

Next, we introduce how to define the state vector.
Position in a poker game is a critical piece of information that determines the order of action.
We define the button (the pre-flop first-hand position), the action position (whose turn it is to take action), and the current game round as one dimension of the state, respectively.
In Poker, the combination of a player's hole cards and board cards determines the game's outcome.
We encode the hole cards and the board cards separately.
The amount of chips is an essential consideration in a player's decision-making process.
We encode this information into two dimensions of the state.
The number of chips in the pot can reflect the action history of both players.
The difference in bets between players in this round affects the choice of action (the game goes to the next round until both players have the same number of chips).
In summary, the state vector has seven dimensions, \ie, button, to\_act, round, hole cards, board cards, chips\_to\_call, and pot.

\section{BigLeduc Poker}
\label{big_leduc_poker}
\begin{figure*}[h]
	\centering 
	\includegraphics[width=1.0\textwidth]{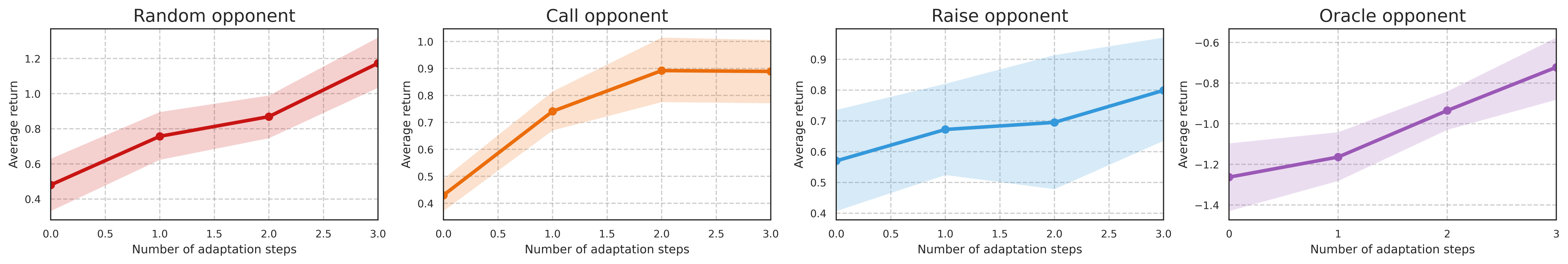}
	\caption{The trained base policy using our L2E framework can quickly adapt to different opponents of different styles and strengths in the BigLeduc poker environment.}
	\label{fast big leduc}
\end{figure*}

We use a larger and more challenging BigLeduc poker environment to further verify the effectiveness of our L2E framework.
The BigLeduc poker has the same rules as Leduc but uses a deck of 24 cards with 12 ranks.
In addition to the larger state space, BigLeduc allows a maximum of 6 instead of 2 raises per round.
As shown in Fig.~\ref{fast big leduc}, L2E still achieves fast adaptation to different opponents.
In comparison with other baseline methods, L2E achieves the highest average return in Table~\ref{bigleduc ave}.
\begin{table}[!htbp]
	\centering
	\caption{The average return of each method when performing rapid adaptation against different opponents in the BigLeduc poker environment.}\label{bigleduc ave}
	\begin{tabular}{ccccc}
		\toprule
		& Random& Call& Raise& Oracle\\
		\midrule
		L2E& \textbf{0.82$\pm$0.28}& \textbf{0.74$\pm$0.22}& \textbf{0.68$\pm$0.09}& \textbf{-1.02$\pm$0.24} \\
		MAML& 0.77$\pm$0.30& 0.17$\pm$0.08& -2.02$\pm$0.99& -1.21$\pm$0.30 \\
		Random& -0.00$\pm$3.08& -0.00$\pm$2.78& -2.83$\pm$5.25& -1.88$\pm$4.28 \\
		TRPO& 0.19$\pm$0.09& 0.10$\pm$0.16& -2.22$\pm$0.71& -1.42$\pm$0.47 \\
		TRPO-P&  0.36$\pm$0.42&  0.23$\pm$0.21&  -2.03$\pm$1.61& -1.69$\pm$0.69 \\
		EOM&  0.56$\pm$0.43&  0.15$\pm$0.13&  -1.15$\pm$0.12&  -1.63$\pm$0.62 \\
		\bottomrule
	\end{tabular}
\end{table}

\section{Grid Soccer}
\label{grid soccer}
This game contains a board with a $6\times9$ grid, two players, and their respective target areas.
The position of the target area is fixed, and the two players appear randomly in their respective areas at the start of the game.
One of the two players randomly has the ball. The goal of all players is to move the ball to the other player's target position.
When the two players move to the same grid, the player with the ball loses the ball. Players gain one point for moving the ball to the opponent's area.
The player can move in all four directions within the grid, and action is invalid when it moves to the boundary.

\begin{figure*}[h] 
	\centering 
	\includegraphics[width=1.0\textwidth]{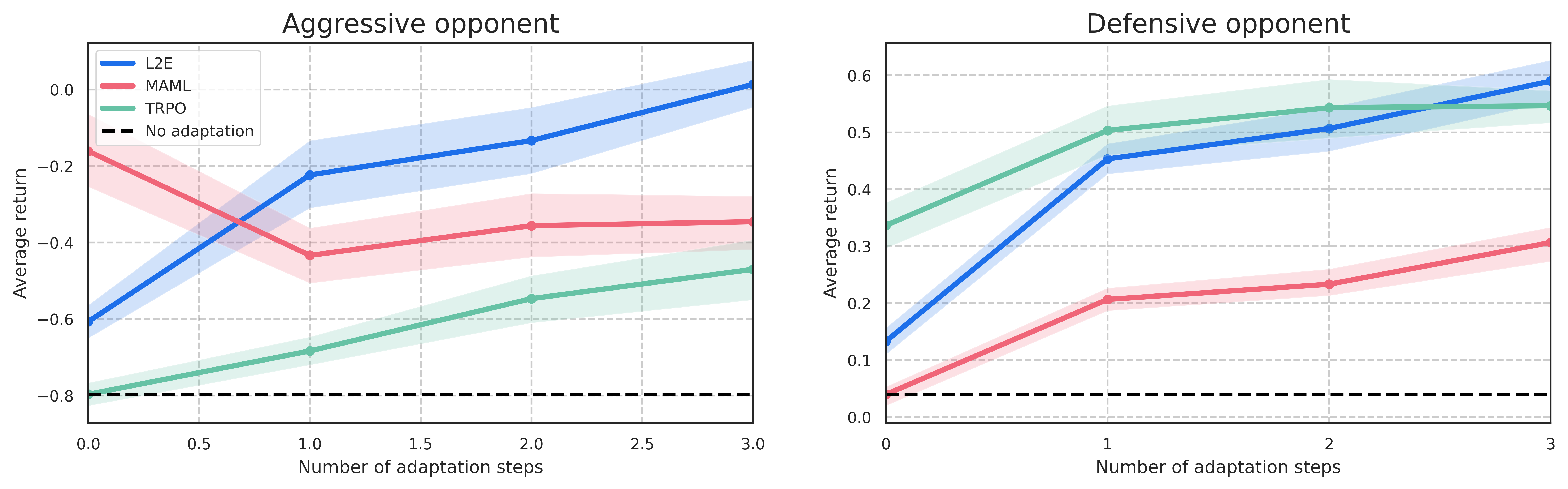}
	\caption{The trained base policy using our L2E framework can quickly adapt to opponents with different styles in a Grid Soccer environment.} 
	\label{soccer exp} 
\end{figure*}

We train the L2E algorithm in this soccer environment in which both players are modeled by a neural network.
Inputs to the network include information about the position of both players, the position of the ball, and the boundary.
We provide two types of opponents to test the effectiveness of the resulting base policy.
1) A defensive opponent who adopts a strategy of not leaving the target area and preventing opposing players from attacking.
2) An aggressive opponent who adopts a strategy of continually stealing the ball and approaching the target area with the ball.
Facing a defensive opponent will not lose points, but the agent must learn to carry the ball and avoid the opponent moving to the target area to score points.
Against an aggressive opponent, the agent must learn to defend at the target area to avoid losing points.
Fig.~\ref{soccer exp} shows the comparisons between L2E, MAML, and TRPO. L2E adapts quickly to both types of opponents;
TRPO works well against defensive opponents but loses many points against aggressive opponents;
MAML is unstable due to its reliance on task specification during the training process.

\newpage
\section{Hyper-parameters}
The hyper-parameters of all experiments are shown in the table below.
\begin{table}[htpb]
	\centering
	\caption{Hyper-parameters of L2E.}
	\scalebox{0.85}{
		\begin{tabular}{cc}
			\hline 
			Hyper-parameter&Value\\
			\hline 
			Training step size hyper parameters $(\alpha,\beta)$ &(0.1,0.01)\\
			Testing step size hyper parameter $\gamma$  & 0.1 \\
			Number of opponents sampled per batch & 40 \\
			Number of trajectories to sample for each opponent & 20 \\
			Number of gradient steps in the training loop & 1 \\
			Number of gradient steps in the testing loop & 3 \\
			Policy network size(Leduc, BigLeduc, Grid Soccer) &  [64,64,(4,4,5)] \\
			Number of training steps required for convergence in Leduc & 300 \\
			Number of training steps required for convergence in BigLeduc & 400 \\
			Number of training steps required for convergence  in GridSoccer & 300 \\
			Hard-to-exploit opponent training epochs & 20\\
			Diverse opponent training epochs  &     50 \\
			Weight of the MMD term $\alpha_\text{MMD}$  &  0.8\\
			Bandwidth of RBF kernel &                   1\\
			Minimum trajectory length N to calculate MMD term & 20 \\
			Number of opponent strategies generated by OSG per round of iteration & N$\le$5 \\
			Number of trajectories sampled to compute MMD term &   8\\
			\hline 
	\end{tabular}}
	\label{l2eparameter}
\end{table}

\end{document}